\documentclass{article}
\usepackage{ijcai21}

\usepackage{times}
\usepackage{soul}
\usepackage{url}
\usepackage[utf8]{inputenc}
\usepackage[small]{caption}
\usepackage{graphicx}
\usepackage{amsmath}
\usepackage{booktabs}
\usepackage{microtype}
\usepackage[hidelinks]{hyperref}

\hypersetup{bookmarksdepth=2}

\newif\ifarxiv
\arxivtrue

\ifarxiv
\usepackage[keeplastbox]{flushend}
\renewcommand{\citeauthor}{\NOTAVAILABLE}
\renewcommand{\citeyear}{\NOTAVAILABLE}
\fi

\title{Where is your place, Visual Place Recognition?}

\author{
Sourav Garg*
\and
Tobias Fischer* {\normalfont (*: equal contributions)}
\And
Michael Milford
\affiliations
QUT Centre for Robotics, Queensland University of Technology
\emails
\{s.garg, tobias.fischer, michael.milford\}@qut.edu.au
}

\ifarxiv
\newcommand{\citemore}[2]{%
    \ifx&#1&%
      \cite{#2}%
    \else%
      \ifx&#2&%
        \cite{#1}%
      \else%
        \cite{#1,#2}%
      \fi
    \fi%
}%
\else
\newcommand{\citemore}[2]{%
  \ifx&#1&%
  \else%
    \cite{#1}%
  \fi%
}%
\fi

\begin{document}

\maketitle

\begin{abstract}
Visual Place Recognition (VPR) is often characterized as being able to recognize the same place despite significant changes in appearance and viewpoint. VPR is a key component of Spatial Artificial Intelligence, enabling robotic platforms and intelligent augmentation platforms such as augmented reality devices to perceive and understand the physical world. In this paper, we observe that there are three ``drivers'' that impose requirements on spatially intelligent agents and thus VPR systems: 1) the particular agent including its sensors and computational resources, 2) the operating environment of this agent, and 3) the specific task that the artificial agent carries out. In this paper, we characterize and survey key works in the VPR area considering those drivers, including their place representation and place matching choices. We also provide a new definition of VPR based on the visual overlap -- akin to spatial view cells in the brain -- that enables us to find similarities and differences to other research areas in the robotics and computer vision fields. We identify numerous open challenges and suggest areas that require more in-depth attention in future works.
\end{abstract}

\section{Introduction}
Visual Place Recognition (VPR) is a rapidly growing topic: Google Scholar lists over 2300 papers matching this exact term, with 1600 of them published since the pivotal survey paper by\ifarxiv\ Lowry et al.~in 2016~\cite{Lowry2016}\else~\citeauthor{Lowry2016}~in \citeyear{Lowry2016}\fi. While exhaustive surveys of works on VPR are given elsewhere~\cite{Lowry2016,zhang2020visual,masone2021survey}, our goal here is to lay a concrete understanding of VPR as a research problem. We argue that research on VPR has increasingly become more dissociated: there is no standard definition of a `place' and comparison of methods is challenging as benchmark datasets and metrics vary substantially. 

\begin{figure}
    \centering
    \includegraphics[width=0.99\linewidth]{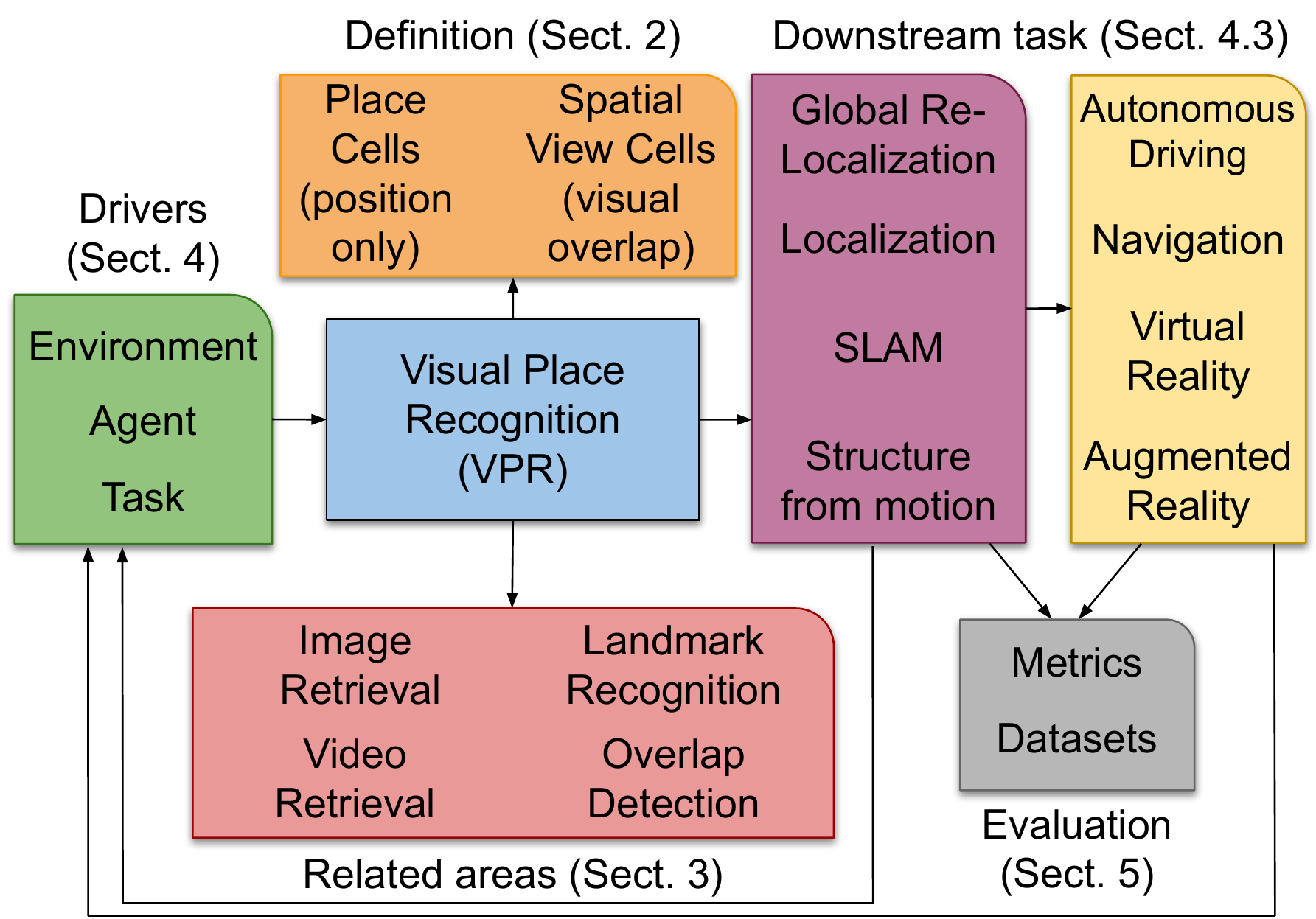}
    \vspace{0.1cm}
    \caption{Visual Place Recognition (VPR) is the ability to recognize one's location based on two observations perceived from overlapping field-of-views. This figure illustrates the main sections of this paper and how they interrelate.}
    \vspace{0.1cm}
    \label{fig:main}
\end{figure}

In light of the dissociation and while retaining the accessibility of a compact treatment, we discuss VPR with regards to its definition (Section~\ref{sec:definition}), how it closely relates to other areas of research (Section~\ref{sec:relatedareas}), what the key drivers of VPR research are (Section~\ref{sec:drivers}), how to evaluate VPR solutions (Section~\ref{sec:evaluation}), and what key research problems still remain unsolved (Section~\ref{sec:openproblems}). Fig.~\ref{fig:main} illustrates the outline of our paper and shows how the various sections are interrelated.

\section{What is Visual Place Recognition?}
\label{sec:definition}

\ifarxiv Lowry et al.~\cite{Lowry2016}~\else\citeauthor{Lowry2016}~\fi state that VPR addresses the question of ``given an image of a place, can a human, animal, or robot decide whether or not this image is of a place it has already seen?''. One can easily see that such a capability is of crucial importance in tasks like localization and navigation, which in turn become ever more important with the advent of artificial intelligence (AI) in autonomous cars, mobile robots that interact with humans, and intelligent augmentation platforms such as the HoloLens 2. Inspiration for VPR is often drawn from the animal kingdom, given the remarkable localization and navigation capabilities of even ``simple'' animals, and the relatively well understood underlying mechanisms (even leading to 2014's Nobel Prize for the discovery of place cells and grid cells, as further detailed below).

Although it might seem inevitable to define a place first, we instead define VPR directly, remembering that it is a comparison of visual data, observed from same or different physical locations with same or different viewpoints. We argue that two such observations can lead to successful recognition if there exists a certain degree of visual overlap due to overlapping field-of-view of the underlying sensor, whereby the acceptable degree depends on the drivers introduced in Section~\ref{sec:drivers}. This implies that: 1) being at the same physical location is not sufficient, the orientation (i.e.~viewpoint) needs to be somewhat similar as well, and 2) places can also be recognized when observed from distant physical locations. In short, we define VPR as the ability to recognize one's location based on two observations perceived from overlapping field-of-views. Note that our definition requires rethinking the typical notion of localization threshold (as used by almost all datasets and evaluation metrics, see Section~\ref{sec:evaluation}), which is based on metric distances without considering orientation.

Our definition is complementary to that of~\ifarxiv Lowry et al.~\cite{Lowry2016}\else\citeauthor{Lowry2016}\fi, but has a different underlying motivation. \ifarxiv Lowry et al.'s\else\citeauthor{Lowry2016}'s\fi~definition is in line with the notion of place cells, which fire when an animal is in a particular place in the environment, irrespective of the animal's orientation. Instead, our definition is in line with spatial view cells, which fire when a specific area of the environment is gazed at by the animal, irrespective of the particular location~\cite{10.1093/cercor/9.3.197}.

We note that in the context of robotics, VPR often involves sequential imagery
rather than single images, as this can significantly improve place recognition performance, especially so for challenging environments~\cite{Lowry2016}. For such sequence-based methods, the equivalent of visual overlap is the overlap of the volume spanned by the sequence.

\section{Related Areas}
\label{sec:relatedareas}
In this section, we highlight similarities and differences of VPR with a handful of related areas. While the relation to image retrieval has been discussed in other works~\cite{Lowry2016,zhang2020visual,masone2021survey}, it is for the first time that VPR's relation to video retrieval, visual landmark recognition, and overlap detection is systematically presented. We argue that for each of those areas, there is a potential for mutual benefits: research into VPR can offer insights for these areas and vice versa. %

\vspace{0.1cm}\noindent\textbf{Image Retrieval:}
Image retrieval refers to the general problem of retrieving relevant images from a large database~\cite{arandjelovic2016netvlad}. VPR is commonly cast as an image retrieval problem that involves a nearest neighbor search of compact global descriptors~\citemore{arandjelovic2016netvlad}{babenko2015aggregating,sivic2003video,Torii2018,jegou2010aggregating} or cross-matching of local descriptors~\citemore{liu2020densernet}{taubner2020lcd,yue2019robust,hausler2021patch,tourani2020early}. With regards to solving the nearest neighbor search problem, VPR and image-retrieval systems face similar challenges. However, the underlying goals differ between the two areas. For image retrieval, similarity criterion could be based on semantic categories such as `clothes' as a product category or nighttime image as an environmental condition category.
However, with the additional context of being a `place' (see Section~\ref{sec:definition}), VPR deviates from the process of merely retrieving a ``similar'' image, which instead is one of the challenges of VPR and referred to as \textit{perceptual aliasing} (see Section~\ref{sec:drivers}). The notion of similarity in VPR is constrained to matching spatial information, where images captured from the \textit{same} place would be considered a true match even if environmental conditions are dissimilar (e.g.~day vs night).

\vspace{0.1cm}\noindent\textbf{Video Retrieval:}
With regards to video retrieval, we observe that sequence-based VPR is typically implemented as a decoupled approach, where single image-based retrieval is followed by sequence score aggregation~\cite{milford2012seqslam}. The recent introduction of explicit sequence-based place representations (where the representation itself describes the sequence), posing VPR as a video retrieval problem, opens up new opportunities to obtain solutions robust to extreme appearance variations~\citemore{garg2021seqnet}{garg2020delta,facil2019condition,arroyo2015towards,neubert2019neurologically,takeda2020dark}.

\vspace{0.1cm}\noindent\textbf{Visual Landmark Recognition and Retrieval:} 
Visual landmark recognition is the classification problem of given an image and a set of images belonging to a large variety of landmarks, deciding to which landmark this image belongs.  Recently, the Google-Landmarks dataset~\cite{weyand2020google} presented a new large-scale \textit{instance-level} recognition and retrieval challenge, with the number of landmarks\footnote{In the context of mobile robotics, the term `landmark' is typically used to indicate \textit{any} specific visual entity in the scene relevant for localization~\cite{luo1992neural,xin2019localizing}.} increased from 30,000 to 200,000 in its second version. This large-scale recognition is an \textit{extreme classification} problem~\cite{choromanska2013extreme}, where existing recognition solutions have relied on retrieval (nearest neighbor search)~\cite{teichmann2019detect}. Google-Landmarks comprises \textit{specific} places (with the semantics of unique proper names) as opposed to general place categories (with the semantics of common names)~\cite{zhou2017places,wu2009visual}. 

In contrast, VPR refers to the ability of distinctively recognizing \textit{any ordinary} place or a region in the 3D world,
thus posing an `extremer' classification problem. 
It remains to be seen how methods developed for landmark recognition and retrieval can be leveraged in the context of VPR -- recent advances include learning to aggregate `relevant' landmarks~\cite{teichmann2019detect}%
, as well as jointly training local and global descriptors~\citemore{cao2020unifying}{yang2020ur2kid,sarlin2019coarse}.

\vspace{0.1cm}\noindent\textbf{Visual Overlap Detection:}
As discussed in Section~\ref{sec:definition}, our definition of VPR is based on an overlapping field of view between the two places that should be matched; thus VPR and the area of visual overlap detection become more closely linked. %
The contrast between defining VPR using visual overlap as opposed to ``positional offset'' impacts the choice of ground truth for both training and evaluation procedures. This contrast has recently been shown to lead to noticeable changes in absolute performance when benchmarking localization algorithms~\cite{pion2020benchmarking}. 

As the ground truth visual overlap might not be available for all datasets, overlap detection measures could be used as a supervision signal~\citemore{}{rau2020imageboxoverlap,chen2020overlapnet} to develop better VPR techniques. A noteworthy recent proposal on overlap detection~\cite{rau2020imageboxoverlap} introduced the `normalized surface overlap' to measure the number of pixels of image A visible in image B (and vice versa). This leads to an asymmetric, but interpretable, measure that can also estimate the relative scale between pairs of images.

\section{What \textit{Drives} VPR Research?}
\label{sec:drivers}
This section outlines the three key drivers of spatially intelligent systems, including intelligent autonomous systems operating in industry and household domains. As drivers, we refer to components that typically impose requirements on the system with regards to a) how the problem should be defined, b) how the solution (in the context of VPR: place representation and matching) should be designed, and c) how these solutions should be evaluated, both in terms of datasets and metrics. The three drivers are the \emph{Environment} where an agent operates (Section~\ref{subsec:environmentconsiderations}), the \emph{Agent} on which the spatially intelligent system is deployed (Section~\ref{subsec:platformconsiderations}), and the \emph{Downstream Task} that is performed (Section~\ref{subsec:taskconsiderations}). In practice, different aspects of each of these drivers are simultaneously at play. We detail why it is crucial to understand the influence of these drivers to design better spatially intelligent systems, in particular in the VPR domain. %

\subsection{Environment}
\label{subsec:environmentconsiderations}

The first driver of VPR research is the operating environment, where research often branches out, as methods that work in certain environment types might cease to do so in other environment types. Differing branches include indoor vs outdoor, suburban vs highway, structured vs open, and human-made vs natural. The operating environment is often tightly coupled with the robotic agent choice (Section~\ref{subsec:platformconsiderations}) -- for example, driverless cars do not operate in office environments, or at least should not.

While the general aim of VPR systems is often stated to be invariance to changes in viewpoint \emph{as well as} changes in appearance (including structural, seasonal, and illumination changes)~\cite{Lowry2016,arandjelovic2016netvlad,garg2018lost,zhang2020visual}, we argue that 1) not all environments/agents require invariance to both viewpoint and appearance (as detailed below), and 2) that there is a trade-off between viewpoint and appearance invariance achievable by current systems (as detailed in Section~\ref{subsec:placerepresentation}). Therefore, knowing the operating environment can provide crucial advantages when deciding how to represent and match places. 

\vspace{0.1cm}\noindent\textbf{Structured Environments and Viewpoint Variations:}
In well-structured environments such as road infrastructure, the extent of 6-DoF viewpoint variations is generally confined, e.g.~for driverless cars on roads, viewpoint varies mostly in the yaw direction~\cite{RobotCarDatasetIJRR}. Similar effects in viewpoint variations can be observed for other platforms too. For example, as soon as aerial vehicles reach a certain height, one can assume a planar homography (``flat world''), simplifying template matching~\citemore{saurer2016homography}{toft2020single,mount2019automatic,tourani2020early}. Planar homographies are also present when mounting the camera at a fixed distance from the surface and pointing towards the surface. In structured indoor environments such as warehouses and offices, aisles and corridors enable Manhattan world assumptions and often simplify Simultaneous Localization And Mapping (SLAM)~\cite{li2018monocular}.

\vspace{0.1cm}\noindent\textbf{Environment-Dependent Appearance Variations:}
Appearance invariance is similarly often constrained when assuming a certain operating environment. However, this kind of invariance is harder to quantify as changes in appearance can originate from a wide range of factors: Examples include changes in the time of the day, seasonal changes, structural changes, and weather changes. Therefore, while viewpoint change could be quantified by the metric shift in translation and rotation, there is no linear scale in the difficulty of appearance invariance. There are even some counter-intuitive examples, where a reference image captured outdoor in the morning might be easier to match to a well-lit nighttime image than to an image captured at noon which has shadows cast on a large area of the image~\cite{corke2013dealing}. %

For different platforms and environments, the requirement of representing and matching places in an appearance-invariant or viewpoint-invariant manner can differ significantly. 
For example, driverless cars typically traverse a well-defined route and could trade-off viewpoint-invariance with appearance-invariance which can be relatively more challenging due to variations in the time of day, season, structural changes including roadworks and differing traffic conditions~\cite{Warburg_2020_CVPR}.
On the other hand, when an autonomous agent is deployed indoors or when considering an unmanned aerial vehicle, its route or maneuvers may not always be constrained, thus requiring viewpoint-invariance more than appearance-invariance.

\vspace{0.1cm}\noindent\textbf{Perceptual Aliasing:}
Another consideration that is tied to the operating environment is the extent of perceptual aliasing. Perceptual aliasing is the problem that two distinct places can look strikingly similar, often more similar than the same place observed under different conditions~\cite{Lowry2016}. For example, indoor environments often contain corridors and hallways that are hard to distinguish. In outdoor settings, different places along a highway or a natural vegetative environment tend to be more perceptually aliased than different places within the man-made urban or suburban dwellings.

\vspace{0.1cm}\noindent\textbf{Dynamic Environments:}
Problems related to the operating environment that -- to our knowledge -- have not yet been addressed in VPR research are sensor dust, reflections (in glass or puddles) and undesired objects close to the camera (e.g.~windscreen wipers). Such conditions are expected in challenging environments like mines and forests, which have come into focus in recent years~\cite{garforth2020lost}. It would be interesting to model the impact of such `noise' explicitly or measure the impact of sensor noise in existing VPR systems. 

\subsection{Agent}
\label{subsec:platformconsiderations}
VPR has widespread applications and is thus deployed on a large variety of robotic platforms, including unmanned ground vehicles and autonomous cars~\cite{doan2019scalable}, unmanned aerial vehicles~\cite{zaffar2019} and unmanned underwater vehicles~\cite{li2015high}. Other platforms where VPR is applied are those tightly coupled to human users such as virtual/augmented reality devices~\cite{sattler2016efficient} and mobile phones~\cite{Torii2018}.

\vspace{0.1cm}\noindent\textbf{Computational Resources:}
A robotic agent typically runs a large number of processes, many of them interacting with each other through tools like the Robot Operating System (ROS)~\cite{ROS}. These processes share limited onboard resources and often require cognitive architectures~\cite{FischerFRAI2018} to interact efficiently. Thus the resources dedicated to the VPR system might be relatively small, and a GPU (that significantly boosts inference times of deep networks) might not be available. Similarly, storage limitations could mean that the reference map of the operating environment (in the form of images, global/local descriptors, point clouds) has to be of reasonable size. Section~\ref{sec:openproblems} discusses some of the open problems in VPR in this context, for example, compact global description, efficient indexing and quantization, and hierarchical place matching pipelines. 
\vspace{0.1cm}\noindent\textbf{Suitable Sensor Suite:}
Depending on the agent and the operating environment, robust VPR solutions can be developed by using additional suitable sensors. For example, event cameras perform exceptionally well when a high dynamic range is required, such as when exiting a dark tunnel and moving into bright sunlight~\cite{fischer2020event}. LIDAR-based systems can perceive the scene's geometry even in the most challenging nighttime conditions, although those systems lack appearance information~\cite{guo2019local}. Using omnidirectional cameras or multi-camera rigs increase the field-of-view and thus the visual overlap, which results in reduced complexities in image matching.

Correct sensor type choice can also aid in tackling specific challenges such as nighttime conditions. Crucially, the sensor capabilities should drive the research regarding what characteristics are required in our learned descriptors. We believe that using novel sensor types such as 3D ultrasonic sensors (e.g.~the Toposens TS3) and sensor fusion~\cite{jacobson2015autonomous} could further improve the robustness of VPR systems.

While some sensors can be a replacement for RGB cameras, another area worthy of more thorough investigation is the use of additional information in the form of prior position or ego-motion. For example, one can assume that autonomous cars are equipped with a GPS sensor. Still, despite the popularity of datasets such as Oxford RobotCar~\cite{RobotCarDatasetIJRR} that contain GPS information, it is rarely used for VPR~\citemore{vysotska2015efficient}{} -- using GPS information in environments where available could refocus research on GPS-denied environments like tunnels or underground mines that have distinct challenges. While there are many examples of GPS-denied environments, almost all mobile robots have some odometry information, but it has only been used in a limited manner for VPR~\cite{pepperell2014all}. %

\subsection{Downstream Task}
\label{subsec:taskconsiderations}
Here, we consider the different tasks that an agent (robotic platform or intelligent augmentation) might perform.

\textbf{Localization, SLAM and Kidnapped Robot Problem:}
Some tasks impact the requirements of VPR systems directly. For example, when VPR is used to provide a coarse estimate of the pose within a 6-DoF localization algorithm~\cite{toft2020long}, the error bounds need to be very tight and the visual overlap between the two places relatively large with sufficient parallax. This is opposed to a scenario where loose error bounds are \mbox{sufficient -- a} rough location estimate might sufficiently narrow down the search space for a subsequent laser-based pose estimation for global re-localization of a mobile robot (``kidnapped robot problem'')~\citemore{jacobson2021localizes}{}.

The requirements with regards to precision and recall are also varying for different scenarios. When using VPR for generating loop closures for SLAM (i.e.~recognizing that a location has been visited previously, so that a globally consistent map can be built), incorrect matches can lead to catastrophic failures, thus requiring high precision VPR~\cite{cadena2016past}. 
On the other hand, one could use VPR to select \mbox{top $k$} matches which are then passed to computationally more intensive stages; in this case, higher recall is more important than the precision. Thus, the downstream task is a key determining factor for formulating and evaluating VPR, as further discussed in Section~\ref{sec:evaluation}. 

We note that a purely \textit{topological} visual SLAM system can be defined through VPR, which is highly relevant for large-scale mapping~\cite{cummins2008fab,doan2019scalable}. Such a topological SLAM system requires determining whether the currently observed place is a revisited one or is a new `unseen' place, thus posing unique design requirements on VPR.

\vspace{0.1cm}\noindent\textbf{Higher-level Tasks:}
The requirements of some downstream tasks like SLAM and Structure from Motion (SfM) are relatively well understood; yet, these requirements are very distinct and probably need a suitably tailored treatment. For example, SfM-based large-scale 3D reconstruction is typically performed offline~\cite{schoenberger2016sfm} and needs sub-pixel accurate alignment of images. The computational requirements of a VPR system then play a much lesser role than in real-time deployments on a mobile platform mapping an unknown environment using visual SLAM.

The requirements of other ``higher-level'' tasks such as those of augmented reality platforms and navigation are not yet well established. This is in part due to the complex hierarchical nature of typical spatially intelligent systems, for example an augmented reality platform would involve many interrelated components such as image retrieval, sequential localization, local feature matching, visual odometry, and pose refinement~\cite{stenborgusing}. %
Furthermore, the utility of VPR and mapping for navigation purposes~\cite{milford2007spatial} is a vastly unexplored area, and a deeper understanding of task requirements is needed.

\section{How to Evaluate Visual Place Recognition?}
\label{sec:evaluation}

This section discusses the evaluation datasets and metrics, in the context of the \textit{drivers}.

\subsection{Evaluation Datasets}
There are numerous place recognition datasets, each covering different aspects of VPR (for recent overviews see~\cite{Warburg_2020_CVPR,masone2021survey}). Thus some datasets are better suited to investigate specific configurations of proposed drivers (i.e.~environments, agents and downstream tasks, see Section~\ref{sec:drivers}), while other datasets better represent different scenarios. This highlights the importance of clearly stating the application scenario targeted by a particular VPR system -- it may be sufficient that the VPR system excels in datasets that are close to the actual use-case (but not in others). 

Recent progress has enabled easier comparison of different methods. VPR-Bench~\cite{zaffar2020vpr} provides a mechanism for the comparison of a new method on an extensive range of datasets. In the light of highly successful standard benchmark datasets in other research areas like visual object tracking~\cite{kristan2018sixth}, we believe that such benchmarking will accelerate VPR research. Mapillary Street Level Sequences (MSLS)~\cite{Warburg_2020_CVPR} is a single dataset that tries to capture all variations of appearance/viewpoint change at once. MSLS notably also introduces `sub-tasks' that can be separately investigated, including sub-tasks like summer to winter, day to night, and old to new. An additional benefit of MSLS is that it provides a hold-out test set that can be used for challenges.

If the aim is to design a VPR system applicable in all different scenarios, an open challenge is to design systems that are equally applicable indoors and outdoors. Few studies evaluate systems both indoors and outdoors, one of them being the above mentioned VPR-Bench~\cite{zaffar2020vpr}. VPR-Bench has shown that performance trends can vary noticeably across environment types, e.g.~indoor vs outdoor. 
However, care should be taken to not make generic assumptions about an architecture when the trained descriptors heavily depend on the training data -- the training data should be representative of the data encountered at deployment time. Most recently,~\cite{Warburg_2020_CVPR} have shown that training on more diverse data drastically improves performance on unseen data. %
This is distinct from the approach where different network configurations are explicitly trained for different scenarios, e.g.~one for indoors and another for outdoors~\cite{sarlin2020superglue}. 

\subsection{Evaluation Metrics}
\label{subsec:evaluationmetrics}
The previous examples show that the downstream task and the relevant evaluation metrics are tightly coupled.
However, we note that many VPR papers do not state why a particular evaluation metric was chosen. Notable exceptions include system papers where VPR is one of many components, and a specific downstream task is considered, such as~\cite{cummins2008fab}. %
The computer vision community typically uses the Recall@K measure, which indicates that in this context the VPR system is benchmarked based on its ability to retrieve a correct match within the top-K retrievals regardless of the false matches. %
On the other hand, the mean average precision (mAP) metric~\citemore{philbin2007object}{}, used in the image retrieval community, explicitly penalizes selection of false matches. The mAP metric could be adopted to measure VPR performance for SLAM-like downstream tasks (Section~\ref{subsec:taskconsiderations}) where precision is more important, complementing measures like Recall at 100\% Precision. %

The area under the precision-recall curve and the F-score 
are sometimes used as summary statistics~\cite{molloy2020intelligent}. However, their practical use is unclear, as these summary statistics imply that recall and precision are of similar importance, which is unlikely the case for most downstream tasks. 
Moreover, these measures are based on the distribution of match scores which may only be relevant for topological SLAM-like scenarios where VPR needs to be highly precise and no subsequent outlier rejection method is employed.

Most of the VPR datasets in robotics are in the form of trajectories with inherent sequential information (Section~\ref{sec:definition}). Thus, evaluation metrics such as `maximum open-loop distance traveled' (that is, the extent of visual odometry or dead reckoning based robot motion without loop closures) have also been considered in the literature~\citemore{clement2020learning}{porav2018adversarial}. We believe it would be beneficial to investigate metrics that tightly couple single-image and sequence-based VPR.

\section{What Are Open Research Problems?}
\label{sec:openproblems}
This section aims to highlight open research problems, considering the drivers discussed in Section~\ref{sec:drivers}. For space reasons and to avoid duplication, we do not cover the open research problems discussed in recent surveys on deep learning methods for VPR~\cite{zhang2020visual,masone2021survey}, which include using autoencoders as an alternative to Convolutional Neural Networks (CNNs), use of generative methods including Generative Adversarial Networks (GANs), using semantic information, making use of heterogeneous data including multi-sensory fusion, and the choice of loss function. 

Here, we broadly classify the open research problems into: 1) representation, discussing the need for better global descriptors and enriched/synthesized reference maps, and 2) matching, discussing the need for better hierarchical matching frameworks, relevant distance metrics and `learning to match'.

\vspace{0.1cm}
\subsection{Place Representation}
\label{subsec:placerepresentation}

\noindent\textbf{Global Descriptors -- Appearance \& Viewpoint Invariance:}
Section~\ref{subsec:environmentconsiderations} discussed the requirements on viewpoint and appearance invariance depending on the operating environment. Here we note that there is a trade-off when learning a descriptor of a fixed size/type: increasing viewpoint-invariance will inevitably reduce some degree of appearance invariance (assuming the same amount of training data)~\citemore{}{arandjelovic2016netvlad,chen2017deep,garg2019semantic}. This is evident from significant differences observed in place recognition performance when considering a cross-combination of datasets such as Nordland (same view, varying appearance)~\citemore{}{sunderhauf2013we} and Pittsburgh (similar appearance, varying view)~\citemore{}{arandjelovic2016netvlad} with feature learning/aggregation methods such as HybridNet (viewpoint-assumed)~\citemore{chen2017deep}{} and NetVLAD (viewpoint-agnostic)~\citemore{arandjelovic2016netvlad}{}. 

There is a vast research gap, and a need for global description mechanisms that go beyond the binary nature of encoding viewpoint, that is, viewpoint-assumed vs viewpoint-invariant. This might be achieved by learning novel ways to incorporate local geometric information in the global descriptor formation, such as using vertical blocks (Stixels)~\cite{hernandez2019slanted}, semantic blobs~\cite{gawel2018x}, objects~\cite{qin2021semantic} or superpixels~\cite{neubert2015superpixel}, where learning could be based on attention mechanisms such as that employed in Transformers~\cite{vaswani2017attention} and Graph Neural Networks~\cite{velivckovic2017graph}. 

\vspace{0.1cm}\noindent\textbf{Global Descriptors -- Efficiency:}
Most of the state-of-the-art global image descriptors are high-dimensional (with dimensions varying from 512~\cite{sarlin2019coarse} to 70,000~\cite{sunderhauf2015performance}). Increasing the descriptor's dimensionality directly leads to increased computational requirements. To improve efficiency, researchers commonly employ dimension-reduction methods such as Principal Component Analysis (PCA), and have also explored quantization~\citemore{jegou2010product}{brandt2010transform,ge2013optimized,sandhawalia2010searching,kalantidis2014locally}, binarization~\citemore{lowry2018lightweight,arroyo2015towards}{jegou2008hamming}, hashing~\citemore{vysotska2017relocalization}{gionis1999similarity,andoni2006near,weiss2009spectral,han2017mild} and efficient indexing~\cite{cao2020unifying} techniques. 

However, there have not been any attempts to learn these efficiency-inducing processes for VPR, particularly considering that retrieving places can include additional information in the form of sequences or odometry. This could be achieved by learning to reduce dimensions~\citemore{mcinnes2018umap}{amid2019trimap,chancan2020hybrid} or to hash~\cite{wang2017survey}, while maintaining the overall structure of the appearance space learnt through the existing global descriptor methods. Existing efficient VPR techniques consider sequential or odometry information in a decoupled manner~\citemore{vysotska2017relocalization}{garg2020fast} but could benefit from jointly considering additional information when optimizing for efficiency.
\vspace{0.1cm}\noindent\textbf{Enriched Reference Maps:}
With the rapid increase in data gathering, more so in the field of autonomous driving, it is high time to consider the use of an enriched reference map, which could be in the form of multiple reference images per location~\cite{churchill2012practice} or semantically-annotated 3D maps~\cite{garg2020semantics}. 
In the simplest case, choosing the best reference set can lead to vast performance improvements. More sophisticated approaches fuse multiple reference sets to achieve even better performance~\cite{churchill2012practice,linegar2015work,molloy2020intelligent}. Multiple reference sets are often used when long-term autonomy is required, as structural changes can be detected and incorporated over time. While significant progress using multiple reference maps has been made in the past, open questions remain around the increased storage and computational requirements when using multiple reference sets. There is clearly a trade-off, and preliminary efforts in that direction~\cite{doan2019scalable} need further attention.

\vspace{0.1cm}\noindent\textbf{View Synthesis:}
To deal with significant viewpoint variations, researchers have also explored matching through multiple synthesized views of the observed places~\citemore{Torii2018}{milford2015sequence}, although the process requires additional computation. However, this can be mitigated by performing view synthesis offline during the mapping traverse and using compact global descriptors with an efficient nearest neighbor search for retrieval. The current bottleneck of automatically generating accurate and relevant views can potentially be addressed by recent advances in volumetric rendering~\cite{mildenhall2020nerf}, performed offline to generate novel views under novel lighting conditions.

\subsection{Place Matching}

\noindent\textbf{Mutually-informed Hierarchical Systems:}
Different downstream tasks can impose very different requirements on the VPR system (see Section~\ref{subsec:taskconsiderations}). For example, a visual SLAM system can be built \textit{with}~\cite{cummins2011appearance} or \textit{without}~\citemore{angeli2009visual}{cummins2008fab} using a geometric verification step based on local feature matching. \ifarxiv\cite{cummins2011appearance}\else\citeauthor{cummins2011appearance}\fi~found this verification to be particularly essential for large datasets. However, it remains an open question how an effective hierarchical system should be designed, where the variables are: the number of complementary VPR techniques fused~\cite{hausler2019multi}, number of stages, and types of unique methods involved, e.g., query expansion~\cite{chum2011total} and keypoint filtering~\cite{garg2018lost}. 

The complementarity and information transfer within different stages in the hierarchy requires an in-depth investigation. This can reveal answers to several overarching questions: should these stages always operate independently or could earlier stages better \textit{inform} subsequent stages (beyond simply providing candidate images); could one selectively apply a subset of the techniques to save computational resources; how such behavior of hierarchical retrieval can be learnt; and in doing so do some of the stages become redundant.

\vspace{0.1cm}\noindent\textbf{Choice of Distance Metric:}
When comparing the global descriptors of two images, one has to choose a suitable distance metric or a similarity measure. Some of the most commonly employed measures include Euclidean~\citemore{arandjelovic2016netvlad}{yu2019spatial,chen2014convolutional,sunderhauf2015place}, cosine~\citemore{sunderhauf2015performance}{garg2018don,garg2018lost}, and Hamming~\citemore{lowry2018lightweight}{arroyo2014fast,arroyo2016fusion,neubert2019neurologically} distance. While some descriptors are better matched using one distance than the other, 
the range of distances distribution is typically relatively narrow, even in non-matching images of a completely different appearance. Therefore a more systematic investigation considering both a theoretical viewpoint and practical performance implications is needed. In particular, important consideration factors include suitability to loss functions (e.g.~max-margin triplet loss)~\citemore{}{arandjelovic2016netvlad,revaud2019learning,garg2021seqnet}, descriptor normalization~\citemore{}{arandjelovic2013all,garg2018don,schubert2020unsupervised}, whitening~\citemore{}{jegou2012negative,arandjelovic2013all}, feature scaling~\citemore{}{beatty1997dimensionality,li2015feature}, and quantization/binarization~\citemore{}{jegou2010product,lowry2018lightweight}.

\vspace{0.1cm}\noindent\textbf{Learning to Match:}
While learning to `describe' (i.e.~local or global descriptors) has been widely explored, there have been limited attempts to learn to `match'. Such matchers can either be learnt through siamese networks~\cite{altwaijry2016learning} or cross-attention based on graph neural networks~\cite{sarlin2020superglue}. The outcome of such matcher could either be a matched reference index or relative 3D pose~\citemore{gridseth2020deepmel}{}. Learning-to-match frameworks for VPR and localization could potentially eradicate the need for sophisticated matching pipelines.

\section{Conclusions}
\label{sec:conclusions}
This survey defines VPR based on the visual overlap of two observations, in line with spatial view cells found in primates. %
This new definition enabled us to discuss how VPR closely relates to other research areas. This paper also identified and detailed three key drivers of Spatial AI: the environment, agent and downstream task. Considering these drivers, we then discussed numerous open research problems that we think are worth addressing in future VPR research.

To date, VPR research has addressed the problems of representing, associating (matching), and searching of \textit{spatial data}, and is a key enabler of Spatial AI. Further advances in VPR research will require unifying the efforts of the artificial intelligence, computer vision, robotics, and machine learning communities, particularly taking into account embodied agents. To achieve this, an in-depth understanding of the problem, research goals and evaluation protocols is necessary, and this paper takes a step in that direction.

\vspace{.1cm}
\noindent\textbf{Acknowledgments:}
This work received funding from the Australian Government, via grant AUSMURIB000001 associated with ONR MURI grant N00014-19-1-2571. The authors acknowledge continued support from the Queensland University of Technology (QUT) through the Centre for Robotics.

\ifarxiv
\bibliographystyle{ieee_fullname}
\small
\bibliography{ijcai21}

\else
\bibliographystyle{renamed}
\small
\bibliography{ijcai21}
\fi
\end{document}